\documentclass[letterpaper, 10 pt, conference]{ieeeconf}

\IEEEoverridecommandlockouts
\usepackage{amsmath, amssymb}

\usepackage{graphics} 
\usepackage{epsfig} 
\usepackage{times} 
\usepackage{amsmath} 
\usepackage{amssymb}  
\usepackage[table]{xcolor}
\usepackage[ruled,linesnumbered, noend]{algorithm2e}
\SetKwRepeat{Do}{do}{while}
\usepackage{adjustbox}
\usepackage{wrapfig}
\usepackage{siunitx}
\usepackage{subcaption}

\usepackage{soul}
\newif\ifshowhl
\showhltrue            
\ifshowhl\else\fi
\usepackage{booktabs}

\usepackage{cite}
\usepackage{comment}
\usepackage{multirow}
\usepackage{graphicx}
\usepackage{wrapfig}
\usepackage{balance}
\usepackage{capt-of}

\newcommand{\ve}[1]{\mathbf{#1}} 
\newcommand{\tve}[1]{\tilde{\mathbf{#1}}} 
\newcommand{\tsl}[1]{\tilde{\mathcal{#1}}} 

\usepackage{color, colortbl}
\definecolor{Gray}{gray}{0.9}

\usepackage{pifont}

\title{Physics-Informed Tactile-based Shape Estimation\\ of Deformable Linear Objects}

\author{Rotem Atari, Stav Ashur and  Avishai Sintov
\thanks{This work was supported by the Israel Science Foundation (grant No. 451/24).}
\thanks{R. Atari, S. Ashur and  A. Sintov are with the School of Mechanical Engineering, Tel-Aviv University, Israel. Corresponding Author: sintov1@tauex.tau.ac.il.}
}

\begin{document}

\maketitle

\begin{abstract}
Accurate shape estimation of Deformable Linear Objects (DLOs) is a fundamental challenge in robotic manipulation, often hindered by visual occlusions and the high-dimensional configuration space of flexible materials. While Force/Torque (F/T) and vision-centric methods have been proposed, they frequently struggle with grasp-site ambiguities and environmental interference. In this letter, we investigate the hypothesis that global shape estimation can be achieved through tactile sensing and proprioception, without any visual perception. We contend that dense, high-resolution tactile imprints capture critical boundary conditions and contact gradients that are sufficient to resolve the high-dimensional configuration of the DLO. To prove this, we introduce a Physics-Informed Vision Transformer (ViT) framework that maps tactile images to a 3D vertex chain. By leveraging self-attention mechanisms and a physics-weighted loss function, our model enforces structural priors to ensure that the predicted geometry adheres to the object's physical manifold. Experimental results demonstrate that our tactile-based approach achieves superior reconstruction accuracy across Euclidean and Chamfer metrics. Furthermore, we demonstrate the efficiency of this modality, achieving model transfer to novel DLO materials and geometries. Our findings suggest that tactile feedback provides a robust and computationally efficient alternative for DLO state estimation in visually degraded or occluded environments.
\end{abstract}


\section{Introduction}
\label{sec:introduction}

Deformable Linear Objects (DLOs), such as cables, wires and tubes, are ubiquitous across industrial, medical and domestic environments \cite{zhu_challenges_2021}. The ability of robots to manipulate these objects effectively is contingent upon accurate 3D shape estimation, which enables complex tasks including shape control, cable insertion and knot tying \cite{sanchez_robotic_2018, Caporali2026}. Shape control is particularly critical in specialized applications like automotive cable routing and assembly. These tasks require deforming a DLO into a specific target configuration, a process that relies heavily on a tightly coupled loop of perception and manipulation planning \cite{yin_modeling_2021}.

To achieve this, the shape of a DLO must be perceived with high fidelity to provide the necessary feedback for control algorithms. Currently, the predominant approach to DLO state estimation relies on external vision sensors, typically RGB-D cameras. Early methods utilized point cloud registration paired with physics-based simulations to enforce physical consistency \cite{tang_state_2017}, while more recent data-driven architectures leverage deep learning and diffusion models to reconstruct 3D geometries from partial visual observations \cite{lv_learning_2023, zhaole_robust_2024, tang_generative_2026}. However, vision-based perception remains fundamentally constrained by line-of-sight requirements and visual clarity. In practical tasks, external cameras frequently suffer from severe hand-object occlusions caused by the robot's own grippers, workspace fixtures, or physical barriers \cite{zhu_challenges_2021}. Furthermore, optical tracking degrades significantly in complex environments with background clutter or dense cable harnesses, where disentangling identical-looking target DLOs poses an persistent computer vision bottleneck \cite{xiang_trackdlo_2023}. While physics-based models can temporarily bridge momentary visual gaps, they fail under complete, sustained loss of sight or heavy clutter. In contrast, tactile-based shape estimation is indispensable in scenarios where visual feedback is entirely unavailable or rendered unreliable, such as routing cables within an automotive chassis during vehicle manufacturing, maneuvering wire harnesses inside dense electronic enclosures, navigating unlit industrial conduits, or when severe hand-object occlusions completely obstruct the view of contact sites and the curve path.

\begin{figure}
    \centering
    \includegraphics[width=\linewidth]{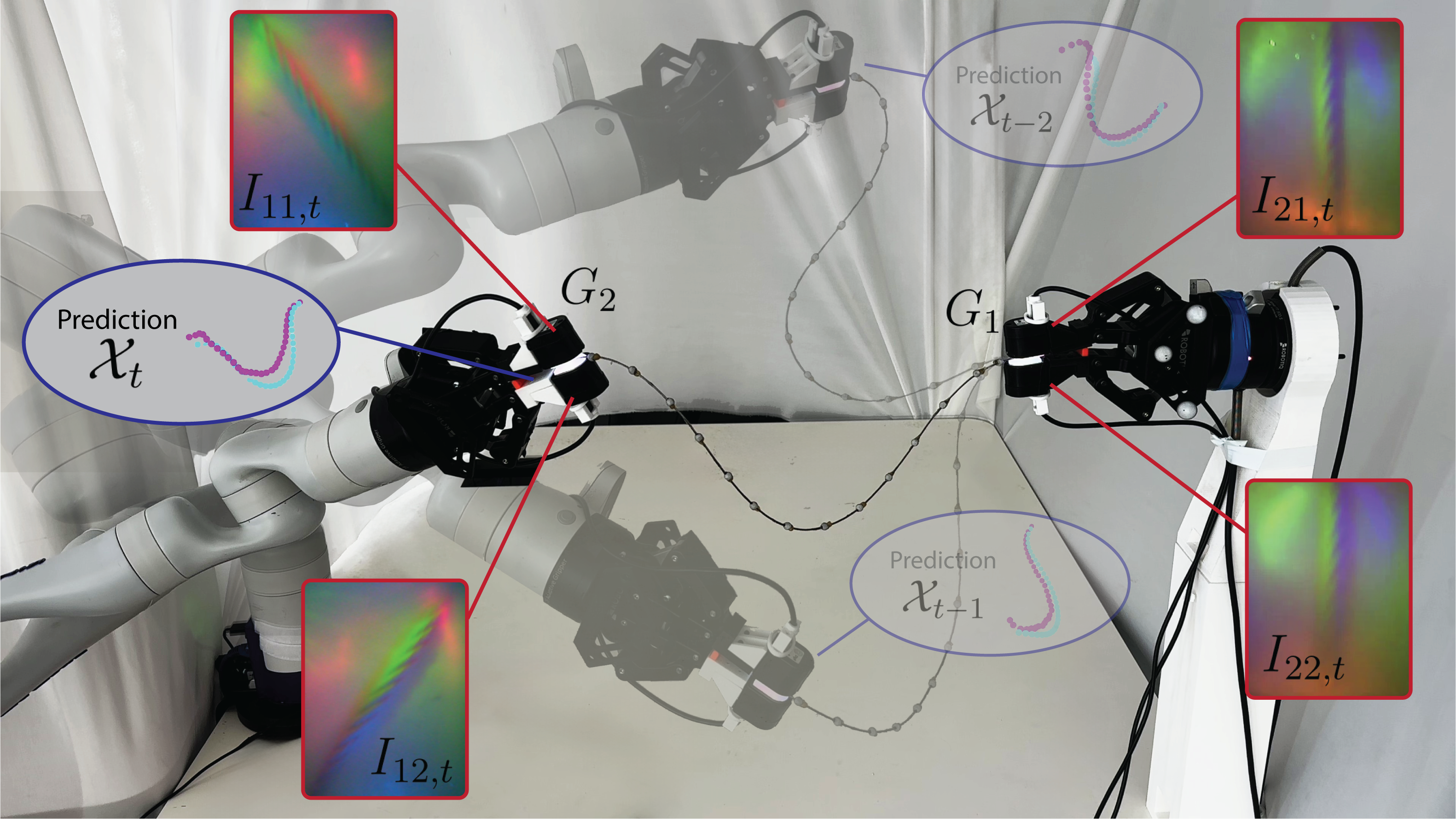}
    \caption{Overview of the tactile-based DLO shape estimation system. Dual robotic grippers ($G_1, G_2$) manipulate a DLO while integrated tactile sensors provide contact images ($I_{ij,t}$).}
    \label{fig:System}
    \vspace{-0.5cm}
\end{figure}

An alternative to vision is the use of force-based sensing for shape estimation. In this domain, significant theoretical groundwork has been laid by Bretl and McCarthy \cite{Bretl2014}, who explored the relationship between contact forces and the equilibrium configurations of elastic rods. Specifically, they have shown that the configuration space of the wire, i.e., the set of all equilibrium configurations, is a six-dimensional smooth manifold. Based on this, later work has proposed a full framework for real-time shape estimation and control of a wire solely using a Force/Torque (F/T) sensor and without any visual feedback \cite{mishani_real-time_2022}. However, translating these force-based approaches into practical robotic applications may be infeasible. These methods are fundamentally based on analytical models that cannot account for the inherent uncertainties in the DLO, such as non-uniform material properties or manufacturing defects. In addition, these models typically assume a perfectly straight shape in the relaxed state and rely on fixed mounting to the robotic arm \cite{Mishani2023}. This rigid setup prevents the system from performing spontaneous pick-and-manipulate tasks, limiting its utility in the unstructured environments where robotic DLO manipulation is most needed. Furthermore, both F/T-based and vision-centric approaches struggle to account for the specific orientation of the DLO tip within the grasp. While F/T methods rely on point-contact assumptions that lack local geometric resolution, visual systems often suffer from severe occlusions at the gripper interface. Consequently, these methods are frequently inapplicable in practical scenarios where the grasp-site orientation varies significantly across trials, as they cannot perceive the critical boundary conditions at the contact level that dictate the global DLO configuration.

Recent advancements in optical-based tactile sensing, such as GelSight \cite{noauthor_gelsight_nodate}, DIGIT \cite{lambeta_digit_2020} and AllSight \cite{azulay2023allsight}, offer high-resolution contact images by embedding miniature cameras behind compliant elastomers \cite{zhang_hardware_2022}. These sensors have been successfully used with cables for local reactive tasks, such as detecting grasp stability \cite{she_cable_2021}, guiding motion primitives for weaving and insertion \cite{wilson_cable_2023}, and performing in-hand DLO following \cite{yu_hand_2024}. Some research has even utilized tactile data to reconstruct the grasped portion of DLOs to enable contour-following \cite{monguzzi_tactile_2023}. However, to the best of the authors' knowledge, optical-based tactile sensing has not yet been utilized to estimate the full 3D global shape of a DLO beyond the immediate contact region. Building upon prior F/T work, we hypothesize that optical-based tactile sensors can capture the same fundamental F/T information while simultaneously encoding the local grasp direction. In this work, we argue that these combined high-resolution features enable the model to resolve the global state of the DLO, effectively moving beyond local reactive control to full-curve shape estimation. 

We propose a novel data-based framework for high-fidelity, vision-free DLO shape estimation. Our primary contributions are as follows. First, we introduce the first framework capable of reconstructing the complete 3D geometry of a DLO using exclusively tactile sensing and proprioception, operating entirely without external visual perception. Second, we introduce a tactile-proprioceptive architecture that leverages the Sparsh tactile foundation model \cite{higuera_sparsh_2024} through a multi-scale attentive fusion mechanism. We employ a cross-attention decoder conditioned on proprioceptive queries, integrated with a differentiable B-spline layer for geometric regularization. This approach demonstrates that the latent representations of Sparsh, originally designed for discrete tasks, can effectively generalize to the structured geometric regression of deformable objects. Consequently, the proposed framework achieves end-to-end, real-time DLO shape estimation directly from raw inputs, entirely bypassing the need for explicit calibration. Third, we formulate a physics-informed weighted loss grounded in the Discrete Elastic Rods formulation \cite{bergou_discrete_2008}, enforcing structural priors on tangency, length and curvature to ensure predicted shapes adhere to the physical manifold of the object. Fourth, we validate our framework through extensive experimentation, including baseline comparisons, ablations of sensor configurations, and transfer learning. Our findings confirm that the model generalizes robustly to DLOs with distinct mechanical and geometric properties using minimal additional data. Lastly, the trained model and labeled dataset is made available open-source for the benefit of the community\footnote{The model, dataset and code will be available upon acceptance for publication.}.

While existing DLO manipulation frameworks rely heavily on vision or force tracking, these modalities are vulnerable to hand-object occlusions and grasp ambiguities. Rather than replacing them, this letter investigates whether high-resolution tactile imprints and proprioception can serve as a standalone, vision-free alternative for global shape estimation. We establish a robust baseline for visually degraded environments, ultimately serving as a complementary modality to fuse with vision pipelines for fail-safe estimation under severe occlusions.

\section{Method}
\label{sec:Method}

\subsection{Problem Formulation}
\label{sec:problem_formulation}

We consider the problem of a DLO grasped by two parallel grippers: one gripper ($G_1$) remains stationary, serving as a fixed base, while the second ($G_2$) is mounted to a robotic arm to facilitate manipulation (Figure \ref{fig:System}). To describe the system's configuration, we define $\mathcal{O}$ as the global coordinate frame, which is rigidly fixed to $G_1$. Each finger of both grippers is equipped with an optical-based tactile sensor that serves as the finger pad. Consequently, let $I_{ij,t}\in\mathcal{I}$ denote the tactile image acquired from sensor $j\in\{1,2\}$ of gripper $G_i$ at time $t$. This configuration yields a comprehensive set of four tactile images at each timestep, defined as $\ve{I}_t=\{I_{11,t},I_{12,t},I_{21,t},I_{22,t}\}$. The pose of $G_2$ with respect to $G_1$ is denoted by $A_t\in SE(3)$, and can easily be computed through the forward kinematics of the arm.

\begin{figure*}
    \centering
    \includegraphics[width=\linewidth]{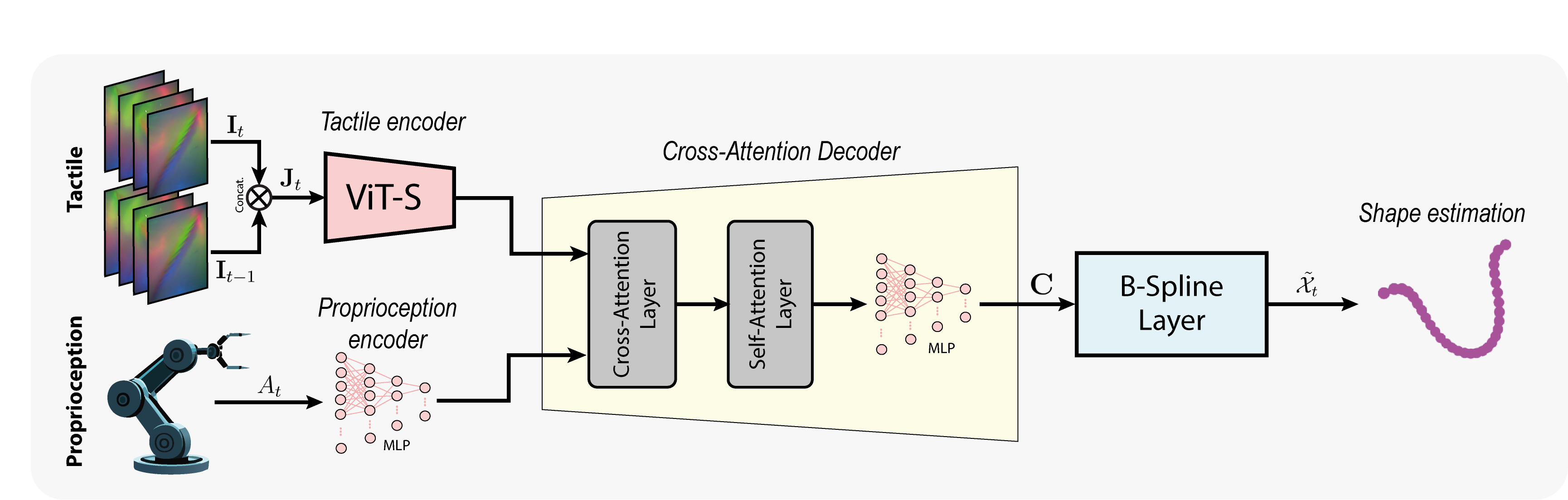}
    \caption{Architecture of the proposed ViT-Sparsh (ViT-S) -based shape estimation model. The system processes synchronous inputs from two grippers, each providing tactile images and gripper pose data. A ViT-based encoder extracts high-dimensional tactile features which are concatenated with proprioceptive embeddings to form input tokens. These tokens are processed by a cross-attention decoder to capture long-range spatial correlations between the two contact sites. Finally, a differentiable B-spline layer enforces geometric smoothness. 
    The output is a spatial vertex chain representing the DLO configuration.}
    \label{fig:model_architecture}
    \vspace{-0.5cm}
\end{figure*}

A DLO is grasped between the two finger pads of each gripper. It is important to emphasize that the DLO tips are oriented arbitrarily with respect to the grippers. This implies that the local contact geometry is not pre-aligned to some axis. The DLO is represented as a discrete ordered set of $n$ vertices $\mathcal{X} = \{\mathbf{x}_1, \mathbf{x}_2, \ldots, \mathbf{x}_{n}\}$, where $\ve{x}_i \in \mathbb{R}^3$ is expressed with respect to coordinate frame $\mathcal{O}$.
We explore the learning of a discrete map 
\begin{equation}
    \Psi_t:\ve{J}_t\times SE(3) \to\mathcal{X}_t, 
\end{equation}
where $\ve{J}_t$ is some product of the tactile space $\mathcal{I}$. Hence, given a tactile measurement $\ve{J}_t$ and the pose $A_t$, the map will output a set of points $\mathcal{X}_t=\Psi(\ve{J}_t, A)$ along the DLO.






\subsection{Data Collection}
\label{sec:data_collection}

To train the proposed model $\Psi$, we utilize an experimental setup to collect a diverse dataset. This data is generated by sampling $N$ DLO configurations, while simultaneously capturing the corresponding tactile images and gripper poses throughout the manipulation process. A motion capture system tracks $m$ reflective markers fixed along the DLO to provide ground-truth data. It should be noted that this system is utilized exclusively for data labeling. During deployment, the model relies solely on tactile sensing and proprioception. Furthermore, while the physical placement of markers provides a relatively coarse resolution of points along the DLO, their individual spatial positions are acquired with high accuracy. Training data is collected in an episodic manner. Prior to the start of each episode, the DLO tips are re-grasped between the grippers in a different configuration, ensuring the DLO is held in various positions and orientations relative to each gripper's coordinate frame. Data collection is performed by driving the robotic arm through random spatial trajectories, subjecting the DLO to various bending and twisting deformations. This process captures the high-dimensional mapping between the DLO's shape and the resulting fusion of tactile imprints and robotic proprioception. We assume a quasi-static regime during manipulation, allowing the model to focus on equilibrium states without accounting for complex inertial dynamics. The generated training data consists of $N$ samples in the form $\mathcal{D}=\{(\ve{I}_i,\mathcal{X}_i, A_i)\}_{i=1}^N$. A similar test dataset was collected in independent episodes for evaluating trained models.


\subsection{State Estimation Model}
\label{sec:state_estimation_model}

To train $\Psi$, we propose a multi-modal encoder--decoder architecture that
fuses tactile image features with robot proprioception to regress
the spatial vertex chain $\mathcal{X}$. The architecture, illustrated in Figure \ref{fig:model_architecture}, comprises four stages: (i) a tactile encoder that
extracts spatial tokens from the four sensors; (ii) a
proprioceptive encoder that maps a $G_2$ pose to a set of
marker queries; (iii) a cross-attention decoder that fuses both
modalities and predicts per-vertex coordinates; and (iv) a
differentiable B-spline layer that enforces geometric smoothness on
the output curve.


\subsubsection{Tactile Encoder}
\label{sec:tactile_encoder}

To enable temporal tactile reasoning, we formulate $\ve{J}_t$ to include temporal tactile images. For each sensor, the three-channel RGB tactile frame $I_{ij,t}$ is concatenated with the previous one $I_{ij,t-1}$. Hence, each tactile sensor provides a six-channel input, encoding local temporal deformation cues. A Sparsh-based Vision Transformer (ViT-S) with a $16 \times 16$ patch size serves as the tactile encoder. This encoder is initialized from the Sparsh foundation model \cite{higuera_sparsh_2024}, which provides a family of general-purpose touch representations. Sparsh was pretrained via DINO self-distillation on a large-scale dataset of over 460,000 tactile images, encompassing a diverse range of vision-based sensors to ensure robust feature extraction. ViT-S remains predominantly frozen to preserve the integrity of the pretrained representations. Task-specific adaptation is facilitated by training only the final two transformer blocks and the concluding LayerNorm at a reduced learning rate, specifically $\times0.05$ the base rate.

Rather than using only the final-layer output, we extract intermediate
representations from a set of the transformer layers
$\mathcal{S} = \{3, 6, 9, 12\}$. They are then combined by a learned weighted sum (attentive probing)
\begin{equation}
\label{eq:multiscale_fusion}
    \mathbf{F}_d
    = \sum_{l \in \mathcal{S}} \alpha_l \,\mathbf{H}^{(l)}~~\text{s.t.}   \qquad
    \alpha_l = \frac{e^{w_l}}{\sum_{l' \in \mathcal{S}} e^{w_{l'}}},
\end{equation}
where $\mathbf{H}^{(l)} \in \mathbb{R}^{N_p \times D}$ represents the layer-normalized output of the $l$-th transformer block. Here, $N_p$ denotes the total number of patch tokens, while $D$ denotes the embedding dimension to be optimized. The terms $w_l$ are learnable scalar parameters optimized during the training process to weigh the contributions of each layer. This mechanism enables the model to leverage both low-level texture features from earlier layers and higher-level semantic features from deeper layers, without increasing the token dimensionality. A single encoder is shared across all four sensors. Then, the four latent representations $F_1,\ldots,F_4$ are concatenated to yield a combined tactile feature map $\mathbf{F}_{\mathrm{tac}} \in \mathbb{R}^{4N_p \times D}$. To disambiguate the spatial origin of each token, a learned per-sensor embedding $\ve{s}_d \in \mathbb{R}^{D}$ is added to the tokens before fusion with the proprioceptive stream.



\subsubsection{Proprioceptive Encoder}
\label{sec:pose_encoder}

The pose $A_t$ is converted into a seven-dimensional vector $\ve{p}_t$, which consists of the spatial position and the orientation expressed as a quaternion for gripper $G_2$. Vector $\ve{p}_t$ 
is mapped through a three-layer Multi-Layer Perceptron (MLP), 
producing an output $\mathbf{z}_p \in \mathbb{R}^{K \cdot D}$.  This vector is reshaped
into $K$ control-point query tokens
$\mathbf{Q} \in \mathbb{R}^{K \times D}$, where $K$ is the number
of B-spline control points. Fixed sinusoidal positional encodings are
added to the queries to encode ordinal positions along the DLO.


\subsubsection{Cross-Attention Decoder}
\label{sec:decoder}

A cross-attention decoder integrates tactile and proprioceptive features using a single cross-attention layer followed by a self-attention layer. Let $\operatorname{MHA}(\ve{Q}, \ve{K}, \ve{V})$ denote the multi-head attention operation, where $\ve{Q}$, $\ve{K}$, and $\ve{V}$ represent the query, key and value matrices, respectively. 
During the cross-attention phase, the control-point queries $\ve{Q}$ attend to the tactile tokens $\ve{F}_{\mathrm{tac}}$ as follows:
\begin{equation}
\label{eq:cross_attn}
    \mathbf{Q}' = \mathbf{Q} + \operatorname{MHA}(\ve{Q},\,
        \ve{F}_{\mathrm{tac}},\,\ve{F}_{\mathrm{tac}}).
\end{equation}
A subsequent self-attention layer with the same configuration allows the
control-point tokens to exchange information along the DLO chain according to
\begin{equation}
\label{eq:self_attn}
    \mathbf{Q}'' = \mathbf{Q}' + \operatorname{MHA}(\mathbf{Q}',\,
        \mathbf{Q}',\,\mathbf{Q}').
\end{equation}
Each refined token is then independently projected to $\mathbb{R}^3$ by a shared MLP, yielding $K$ predicted control points
${\mathbf{C}} = \{{\mathbf{c}}_1, \ldots, {\mathbf{c}}_K\}
\in \mathbb{R}^{K \times 3}$.


\subsubsection{Differentiable B-Spline Layer}
\label{sec:bspline}

To enforce geometric smoothness on the output curve, the predicted
control points in ${\mathbf{C}}$ are passed through a differentiable
cubic B-spline layer. A uniform knot vector is defined over the parameter domain $u \in [-1, 1]$, from which $n$ evaluation points are sampled at equally spaced parameter values $\{u_j\}_{j=1}^{n}$. The final vertex positions are obtained by evaluating the cubic B-spline basis functions $B_{i,3}$ (of degree 3) at each parameter value $u_j$ as follows:
\begin{equation}
\label{eq:bspline}
    \tve{x}_j
    = \sum_{i=1}^{K} \mathbf{c}_i\, B_{i,3}(u_j),
    \qquad j = 1,\ldots,n.
\end{equation}
This yields the predicted vertex set $\tsl{X}_t=\{\tve{x}_1,\ldots,\tve{x}_n\}$ along the DLO. The B-spline layer is parameter-free and fully differentiable, allowing gradients from a loss function to propagate back through the control points. By construction, the B-spline formulation ensures a $C^2$-continuous curve, effectively suppressing high-frequency oscillations that a discrete curvature regularizer might fail to eliminate. Furthermore, this approach decouples the network’s output dimensionality ($K$ control points) from the final evaluation resolution ($m$ sampled points), allowing for a high-fidelity curve representation without increasing the complexity of the prediction head.


\subsection{Loss Function}
\label{sec:loss_function}

The training objective is designed to ensure that the predicted DLO shapes are not only geometrically accurate but also physically plausible. Rather than relying solely on raw vertex positions, we utilize a physics-informed weighted loss alongside a correspondence-free shape similarity term:
\begin{equation}
\label{eq:total_loss}
    \mathcal{L} 
    = \lambda_{\text{w}} \,\mathcal{L}_{\text{weighted}}(\tsl{X}_t, \mathcal{X}_t)
    + \lambda_{\text{c}} \,\mathcal{L}_{\text{chamfer}}(\tsl{X}_t, \mathcal{X}_t),
\end{equation}
with scalar weights $\lambda_{\text{w}}$ and $\lambda_{\text{c}}$. Physics-informed weighted loss $\mathcal{L}_{\text{weighted}}$ and Chamfer distance loss $\mathcal{L}_{\text{chamfer}}$ are presented next.



\subsubsection{Physics-Informed Weighted Loss}
\label{sec:weighted_loss}

To regularize the network's predictions, we exploit the geometric structure of discrete curves to impose constraints derived from the Discrete Elastic Rods (DER) formulation \cite{bergou_discrete_2008}. This approach is motivated by the physical properties of everyday DLOs (e.g., steel wire, USB cables), which are characterized by inextensibility and resistance to bending. By supervising the derivatives of the curve (i.e., tangents, lengths and curvature), we encourage the model to produce smooth, consistently directed shapes and suppress physically implausible high-frequency oscillations that standard positional losses often fail to eliminate.

Let edge vectors $\ve{e}_i = \ve{x}_{i+1} - \ve{x}_i$ with $i=1,\ldots,n{-}1$ connect consecutive vertices, from which we derive edge length $\ell_i = \|\ve{e}_i\|_2$, tangent unit vector $\ve{v}_i = \frac{\ve{e}_i}{\ell_i}$, and the discrete curvature 
binormal \cite{bergou_discrete_2008}
\begin{equation}
\label{eq:curvature_binormal}
    \boldsymbol{\kappa}_i 
    = \frac{2\,(\mathbf{e}_{i-1} \times \mathbf{e}_{i})}
           {\ell_{i-1}\,\ell_{i} + \mathbf{e}_{i-1} \cdot \mathbf{e}_{i}},
    \quad i = 2,\ldots, n{-}1.
\end{equation}
These geometric quantities serve dual roles: as supervision targets 
within the physics-informed loss and as 
differentiable regularizers that enforce inextensibility, local tangent 
consistency and physically plausible bending. Hence, we compute these values given the predicted $\tsl{X}$ and target $\mathcal{X}$ vertex sets. The weighted loss is then
\begin{equation}
\label{eq:weighted_loss}
    \mathcal{L}_{\text{weighted}} 
    = w_n \,\mathcal{L}_{\text{pos}}
    + w_e \,\mathcal{L}_{\text{tan}}
    + w_\ell \,\mathcal{L}_{\text{len}}
    + w_\kappa \,\mathcal{L}_{\text{curv}},
\end{equation}
where $w_n$, $w_e$, $w_\ell$ and $w_\kappa$ are scalar weights of the Mean-Squared Error (MSE) components between the corresponding predicted and ground-truth quantities:
\begin{align}
\label{eq:loss_components}
    \mathcal{L}_{\text{pos}}  &= \frac{1}{n}\sum_{i=1}^{n} 
        \|\tve{x}_i - \ve{x}_i\|_2^2, \\
    \mathcal{L}_{\text{tan}}  &= \frac{1}{n{-}1}\sum_{i=1}^{n-1} 
        \|\tve{v}_i - \ve{v}_{i}\|_2^2, \\
    \mathcal{L}_{\text{len}}  &= \frac{1}{n{-}1}\sum_{i=1}^{n-1} 
        (\tilde{\ell}_i - \ell_i)^2, \\
    \mathcal{L}_{\text{curv}} &= \frac{1}{n{-}2}\sum_{i=2}^{n-1} 
        \|\tilde{\boldsymbol{\kappa}}_i 
        - \boldsymbol{\kappa}_i\|_2^2.
\end{align}
Each component of the weighted loss serves a specific role in ensuring the predicted curve is both geometrically and physically sound. The position term $\mathcal{L}_{\text{pos}}$ provides foundational geometric supervision, while the tangent term $\mathcal{L}_{\text{tan}}$ penalizes deviations in local orientation to encourage smooth, consistently directed curves, even in cases where absolute positional error is minimal. The length term $\mathcal{L}_{\text{len}}$ enforces inextensibility by ensuring that predicted edge lengths match the ground truth. Finally, the curvature term $\mathcal{L}_{\text{curv}}$ regularizes bending through the discrete curvature binormal, effectively suppressing physically implausible high-frequency oscillations that might otherwise appear in the predicted curve.

\subsubsection{Chamfer Distance}
\label{sec:chamfer}

While the ordered terms in $\mathcal{L}_{\text{weighted}}$ rely on a consistent vertex indexing, guaranteed in our pipeline by spline-based resampling, the inclusion of the symmetric Chamfer distance serves as a safeguard against residual resampling ambiguities. By providing a permutation-invariant shape matching signal, it offers a topology-agnostic gradient that focuses on global shape alignment without requiring fixed point-to-point correspondences. This ensures that even if the discrete indexing shifts slightly, the model remains penalized for overall deviations in the DLO's spatial configuration. The symmetric Chamfer distance is defined as
\begin{equation}
\label{eq:chamfer}
\mathcal{L}_{\text{chamfer}}
= \frac{1}{2n}\biggl(\sum_{i=1}^{n} \min_{j} 
            \|\tve{x}_i - \mathbf{x}_j\|_2
      + \sum_{j=1}^{n} \min_{i} 
            \|\mathbf{x}_j - \tve{x}_i\|_2\biggl),
\end{equation}
summing the distances between each point in one set and its nearest neighbor in the other. This complementary loss term enhances the robustness of the training process by providing a signal that is less sensitive to the specific discretization of the DLO.

\section{Experiments}
\label{sec:experiments}

In this section, we evaluate the proposed approach and provide a comprehensive performance analysis across several key metrics. To ensure computational efficiency, all models were trained on an NVIDIA GeForce RTX 5090 GPU with mixed-precision training. 
Videos of 4experiments can be seen in the supplementary material.


\subsection{Hardware Platform}
\label{sec:hardware}

We base our experiments on four DIGIT tactile sensors \cite{lambeta_digit_2020}, mounted on the fingers of two Robotiq 2F-85 parallel grippers. One gripper ($G_1$) is fixed to the workspace while the other ($G_2$) is mounted to a Kinova Gen3 7-DOF manipulator. The grippers hold the two opposite ends of a DLO. The system is seen in Figure \ref{fig:System}. Each DIGIT sensor provides a $320{\times}240$ RGB tactile image at a frequency of approximately $15\,\text{Hz}$. For data collection and to acquire the ground-truth shape of the DLO, an OptiTrack motion capture system tracks $m=14$ reflective markers affixed along the DLO at roughly equal arc-length intervals. The pose $A_t$ of $G_2$ with respect to $G_1$ is acquired through forward kinematics of the robotic arm. Robot motion and data acquisition are controlled using ROS\,2. 



\subsection{Dataset}
\label{sec:dataset_exp}

We begin by collecting data from a primary DLO, a braided steel wire with a diameter of \SI{1.5}{mm} and a length of \SI{550}{mm}. The collection was performed over 36 episodes, where in each episode, the wire was grasped in a new pose somewhat covering the orientation range $[-45^{\circ},45^{\circ}]$ with respect to the gripper axis to ensure grasp diversity. This range represents the physical limit of the setup, as wider angles cause structural collisions or slippage. Then, the robotic arm executed quasi-static gripper motions while deforming the wire under gravity. The $m=14$ unordered tracked markers are sorted via a momentum-based greedy algorithm and resampled to $K=16$ evenly-spaced vertices through B-spline fitting. The value for $K$ was empirically optimized through an ablation study, with the estimation error showing minimal variation across different values. The collection yielded a total of $N=196,678$ labeled training samples. An additional $21,853$ samples were collected over four independent episodes and used as the test set.

\subsection{Model Training}
\label{sec:training_protocol}

All models are trained using the AdamW optimizer ($\beta_1=0.9$, $\beta_2=0.999$) and a OneCycleLR schedule with a $10\%$ linear warmup. The base learning rate and weight decay are both set to $10^{-4}$. To facilitate task-specific adaptation while preserving the Sparsh foundation model's features, the backbone is kept mostly frozen; only the last two transformer blocks are unfrozen and trained at a reduced learning rate of $10^{-5}$. Training is conducted with a batch size of $32$ over $10$ epochs. To improve feature extraction, a reference tactile image (captured under no-contact conditions) is subtracted from the observed tactile frames. This subtraction ensures the model remains agnostic to the sensor's static background and focuses exclusively on the color gradients generated by physical contact. Both the input poses and output marker positions are standardized to zero mean and unit variance to ensure numerical stability and balanced gradient updates. The training objective is the physics-informed loss $\mathcal{L}$ defined in Eq.~\eqref{eq:total_loss}, with the loss hyperparameters set to $\lambda_w = 1.0$, $\lambda_c = 0.25$, $w_n=1.0$, $w_e=0.05$, $w_\ell=0.5$ and $w_\kappa=0.01$. These weights were determined empirically during training to properly balance spatial localization accuracy with physical smoothness constraints, ensuring that the predicted 3D vertex chain adheres closely to the DLO's material continuity without causing optimization instability.


\subsection{Model Evaluation}
\label{sec:model_comparison}

\begin{table*}
  \centering
  \caption{Comparison of shape estimation performance}
  \label{tab:model_results}
  \renewcommand{\arraystretch}{1.15}
  \resizebox{0.8\textwidth}{!}{%
  \begin{tabular}{@{}l c c c c c c}
    \toprule
    \multirow{2}{*}{Model}
        & Num. trainable & Num. total & RMSE & Chamfer & Inference & Freq.\\
       & param. ($\times10^{6}$) & param. ($\times10^{6}$) & (mm) & dis. (mm) & time (ms) & (Hz)\\
    \midrule
    Pose-Only (MLP) & 3.04 & 3.04 & $55.00 \pm 22.00$ & $44.00 \pm 19.00$ & \cellcolor[HTML]{C0C0C0}0.9 & \cellcolor[HTML]{C0C0C0}$1.11\times10^3$ \\
    Pose + Direction & 2.70 & 2.70 & $50.63 \pm 24.25$ & $40.70 \pm 17.06$ & - & -\\
    ResNet-18 + Cross-Attn & 3.49 & 3.49 & $36.00 \pm 20.00$ & $30.00 \pm 16.00$ & 17.6 & 56.82 \\
    Propose model w/o B-spline layer & 5.33 & 23.74 & $35.01 \pm 16.62$ & $30.06 \pm 13.64$ & 61.5 & 16.26 \\
    Proposed model w/o $\mathcal{L}_{\text{weighted}}$ & 6.38 & 24.80 & $37.00 \pm 20.90$ & $30.82 \pm 16.42$ &  61.5 & 16.26\\
    Proposed model & 6.38 & 24.80 & \cellcolor[HTML]{C0C0C0}$30.52 \pm 15.22$ & \cellcolor[HTML]{C0C0C0}$27.29 \pm 12.30$ & 61.5 & 16.26 \\
    \bottomrule
  \end{tabular}%
  }
  \vspace{-0.3cm}
\end{table*}

Standalone analytical estimators (e.g., DER) are not viable baselines here because our arbitrary grasp orientations leave the required boundary conditions unobservable. Furthermore, prior learning-based DLO trackers rely on base F/T sensing rather than distributed tactile arrays, precluding direct deployment. We therefore validate our proposed DLO shape estimation model against five baseline architectures designed to isolate the contributions of different sensing modalities and architectures. The first, a Pose-Only (MLP) baseline, serves as a lower bound by mapping the 7-D gripper pose $A_t$ directly to the predicted vertex chain, attempting to capture only robot proprioception. To investigate the specific tactile features learned by our model, we introduce a pose and DLO tip direction baseline; this model removes complex tactile features in favor of a geometric cue available at the contact site. For each sensor, the 2D tip direction is extracted with respect to the gripper. The direction is derived via a non-differentiable pipeline of adaptive thresholding, morphological cleanup, and robust line fitting to the contact pixels. The direction, in the form of unit vector, is then fed to an MLP alongside the gripper pose to determine if the model primarily reasons about orientation or more complex contact forces. Next, we compare against a ResNet-18 with Cross-Attention architecture, which utilizes a randomly initialized ResNet-18 to encode tactile images into spatial tokens for fusion with proprioceptive data, testing whether generic visual features are sufficient for accurate tactile state estimation in place of our specialized backbone. Finally, we consider the proposed model without the use of the physics-informed weighted loss $\mathcal{L}_{\text{weighted}}$ as well as without the differentiable B-spline layer to analyze their respective contributions.

\begin{figure}
  \centering
  \begin{tabular}{c}
       \includegraphics[width=\columnwidth]{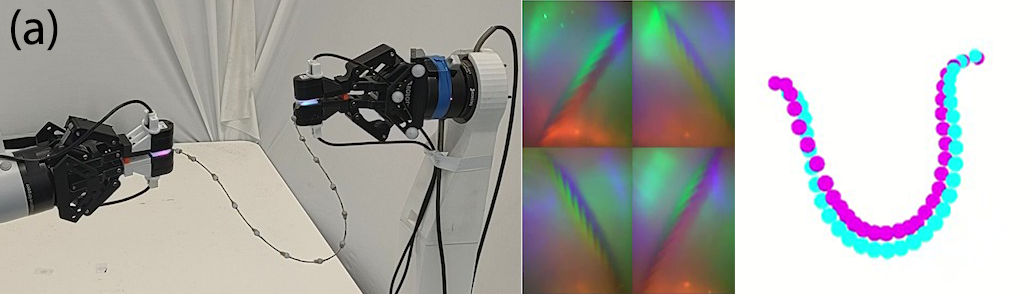} \\
       \includegraphics[width=\columnwidth]{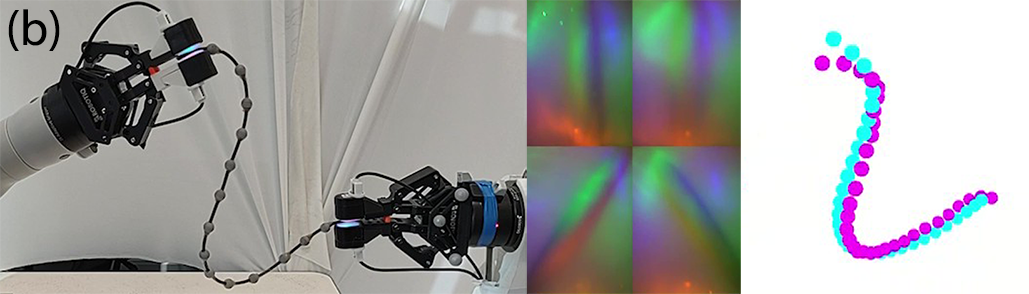} \\
       \includegraphics[width=\columnwidth]{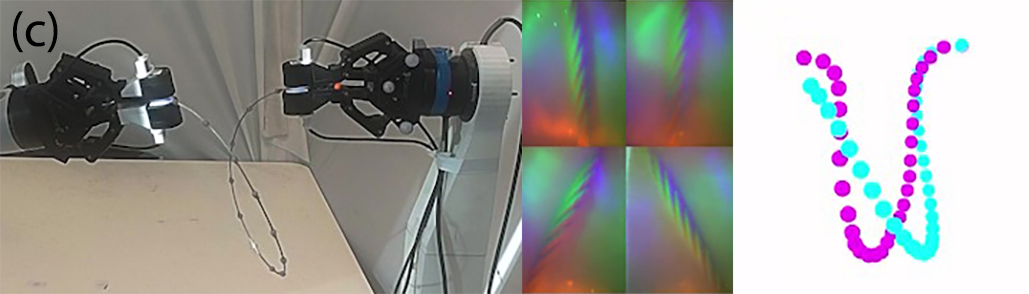} 
  \end{tabular}
  \caption{Visual examples of ground truth and predicted shapes for the three DLOs. (a) 1.5 mm diameter steel cable with estimation error of \SI{34.1}{mm}, (b) 4.5 mm diameter USB cable with estimation error of \SI{17.7}{mm}, and (c) 2.0 mm diameter steel cable with estimation error of \SI{41.5}{mm}. Each row displays the experimental configuration, the corresponding four-channel raw tactile input, and the resulting 3D shape estimation (cyan: ground truth, magenta: prediction).}
  \label{fig:Cable_examples}
  \vspace{-0.2cm}
\end{figure}
\begin{figure}
  \centering
  \includegraphics[width=\columnwidth]{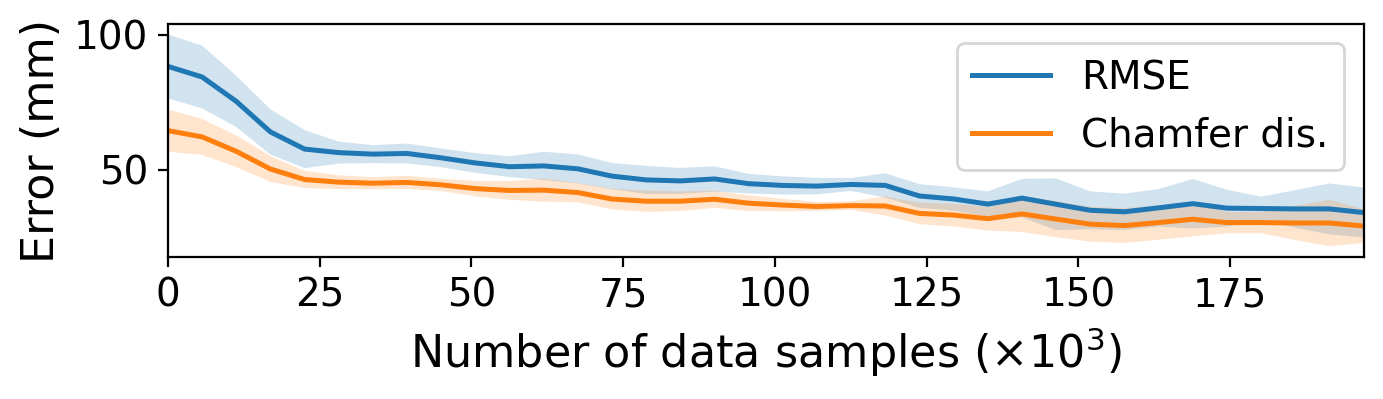}
  \vspace{-0.6cm}
  \caption{Shape estimation accuracy with respect to the number of samples in the training data.}
  \label{fig:data_efficiency}
  \vspace{-0.6cm}
\end{figure}
\begin{figure*}
  \centering
  \begin{tabular}{cc}
      \includegraphics[width=\columnwidth]{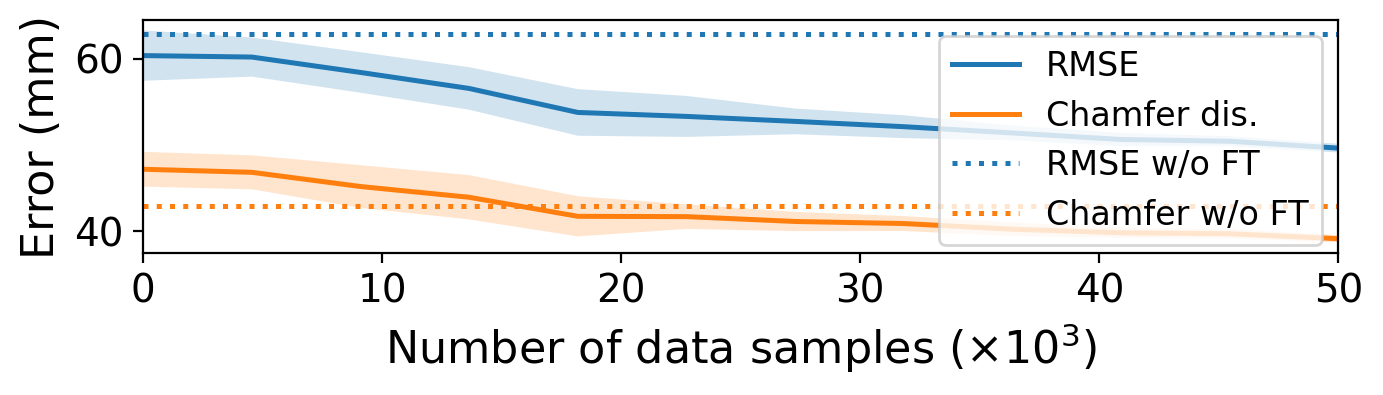} & \includegraphics[width=\columnwidth]{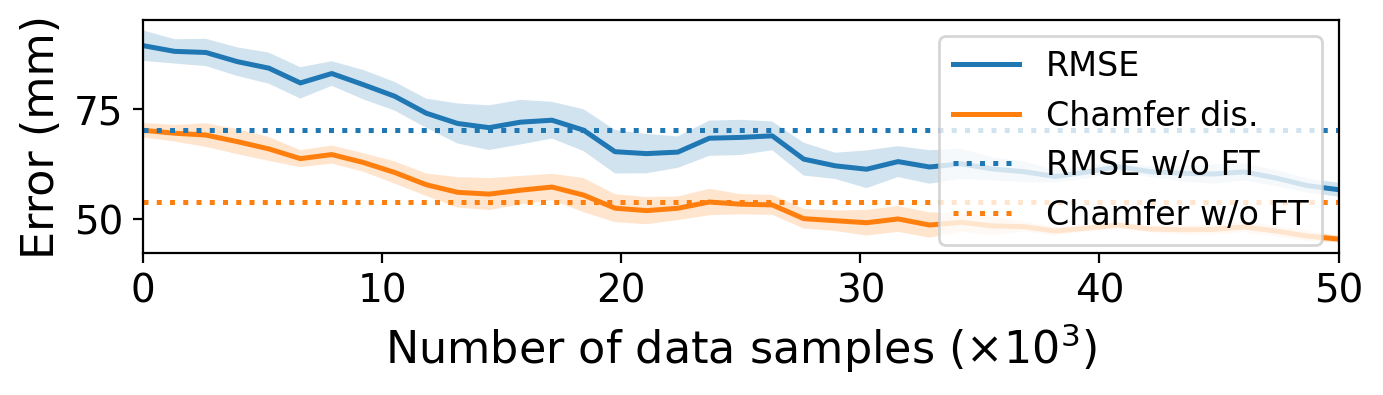} \vspace{-0.2cm}\\
      (a) & (b)
  \end{tabular} 
  \vspace{-0.2cm}
  \caption{Transfer estimation errors for (a) a USB cable of \SI{4.5}{mm} diameter and length \SI{550}{mm} and (b) a steel cable of \SI{2}{mm} diameter and \SI{750}{mm} length, with regard to the number of new training samples used to fine-tune the model. Dotted lines show training of the models from scratch with the new data.}
  \label{fig:Transfer_USB}
  \vspace{-0.5cm}
\end{figure*}

The estimation accuracy of the evaluated models is quantified using two primary metrics: per-vertex Root Mean Square Error (RMSE) between predicted shape vertices and ground-truth coordinates, and the Chamfer distance as defined in \eqref{eq:chamfer}. Table \ref{tab:model_results} presents the results along with the number of parameters in each model. As shown, shape estimation based solely on the gripper poses yields poor predictive accuracy, with higher error margins than the multi-modal variants. This result is consistent with foundational DLO theory, which establishes that a single gripper pose can correspond to a multitude of physically valid configurations \cite{Bretl2014}. Consequently, these findings confirm that proprioception alone is insufficient for robust shape estimation. 

The incorporation of the DLO tip direction at the grasp contacts acts as a beneficial geometric prior, leading to a marginal reduction in estimation error. However, the significant performance gap relative to models utilizing full tactile images indicates that dense, learned tactile features capture critical information beyond simple contact orientation. These features likely encompass deformation gradients, contact geometry and force distributions, which provide the necessary depth to significantly enhance the overall estimation accuracy. These data-driven advantages are further bolstered by our training objective. The experimental results obtained without the physics-informed weighted loss underscore its critical importance, as the absence of these constraints leads to a notable degradation in reconstruction fidelity. This suggests that the $\mathcal{L}_{\text{weighted}}$ term effectively bridges the gap between raw data-driven estimation and the underlying physical manifold of the DLO, ensuring that the model adheres to essential geometric and structural priors during inference. Similarly, removing the B-spline layer eliminates the inherent spatial coupling between neighboring nodes, resulting in high-frequency, physically implausible oscillations and decreased accuracy. Consequently, our proposed ViT-based model achieved superior performance, yielding the lowest estimation error across both Euclidean and Chamfer metrics. This success also suggests that the attention mechanisms in the ViT architecture are particularly well-suited for capturing long-range spatial correlations between local tactile contact gradients and the global DLO configuration. By leveraging these high-dimensional features, the model effectively resolves the inherent ambiguities of proprioception-only approaches, providing a robust and high-fidelity reconstruction of the entire vertex chain. Figure \ref{fig:Cable_examples}a shows a shape estimation instance of the steel DLO. 

\begin{figure}
  \centering
  \includegraphics[width=\linewidth]{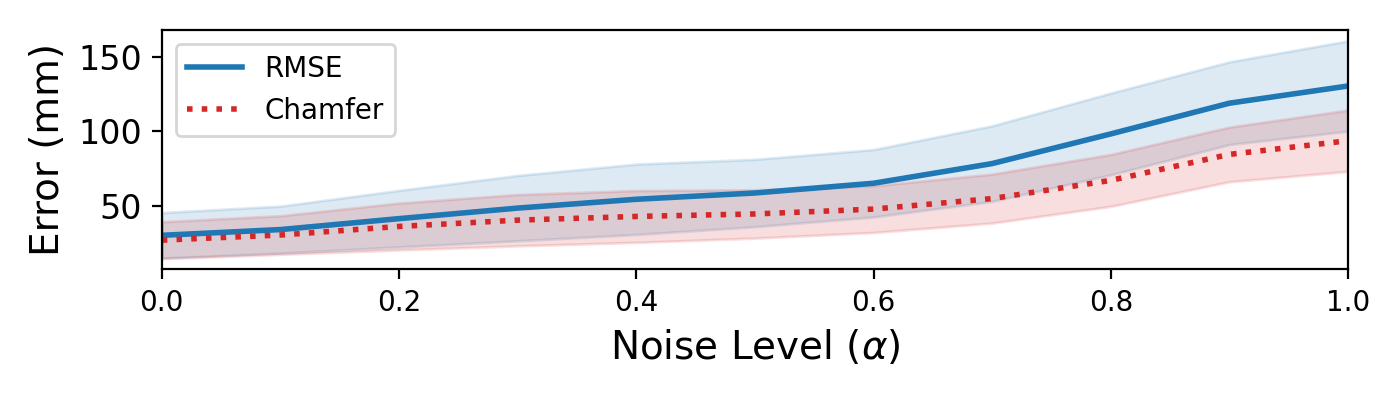}
  \vspace{-0.6cm}
  \caption{Chamfer distance and RMSE for shape estimation with induced noise to the tactile images.}
  \label{fig:noise_robustness_plot}
  \vspace{-0.6cm}
\end{figure}
To evaluate reliability under sensor degradation, we added zero-mean Gaussian white noise to all tactile images in the test set, parameterized by an intensity noise factor $\alpha \in [0, 1]$, where 0 and 1 values correspond to original and fully corrupted images, respectively. As shown in Figure \ref{fig:noise_robustness_plot}, both RMSE and Chamfer distance errors degrade moderately across the noise spectrum. These results demonstrate the framework's robustness, maintaining bounded estimation stability even under severe image corruption.

To assess the computational efficiency and suitability for real-time robotic control, Table \ref{tab:model_results} also reports the mean inference time for each architecture over 1,000 trials. We note that the tip direction computation was not optimized and was utilized solely for accuracy estimation. Consequently, its computational time is not reported, as it does not reflect the potential real-time performance of a production-ready implementation. The results indicate that our proposed ViT-based model exhibits the highest inference time, which is attributed to its larger number of trainable parameters compared to the baseline architectures. Nevertheless, the model achieves a mean inference rate of 16~Hz (61.5~ms latency), matching or exceeding state-of-the-art DLO state estimation pipelines (10-15~Hz) \cite{lv_learning_2023, gu_learning_2025}. Crucially, this execution speed easily surpasses typical closed-loop DLO control frequencies (5-15~Hz) \cite{yu_hand_2024}, ensuring real-time feedback aligned with both sensor throughput and control demands.

\subsection{Data Requirement Analysis}
\label{sec:data_efficiency}

We subsequently analyze the data requirements for training the proposed ViT-based architecture. To quantify how the size of training data influences performance, we evaluate the model's accuracy across a varying number of collection episodes. This evaluation is conducted via a cross-validation approach: for each specified episode count, the model is trained 10 independent times using randomly sampled episodes from the full dataset. Each episode contains approximately $5,460$ training samples, allowing us to characterize the learning curve and identify the point of diminishing returns for data collection. 

Figure \ref{fig:data_efficiency} illustrates the evolution of the RMSE and Chamfer distance as a function of the number of training episodes. Both metrics exhibit a sharp initial improvement after the first five episodes, with the RMSE decreasing by 35\% and the Chamfer distance by 28\%. This steep trajectory suggests that the model rapidly internalizes the behavior of DLO deformation, while the subsequent leveling of the curves indicates the onset of diminishing returns as the dataset grows beyond a critical threshold. Performance continues to improve gradually as more data is introduced, though at a significantly reduced rate compared to the initial phase. Notably, the inter-trial variance increases with more than $25$ episodes, suggesting that in higher data regimes, the specific composition of episodes becomes more influential. This phenomenon is likely due to the inherent variation in DLO configurations across different recording episodes, where certain rare geometries may disproportionately affect model convergence. Overall, these results indicate that reliable state estimation is achievable with as few as $15$ episodes (${\sim}42\%$ of the full dataset). This has significant practical implications, as it demonstrates that data collection effort can be substantially reduced in new deployment scenarios without compromising the system's operational integrity. Hence, we next investigate the ability to transfer the model to unseen DLOs.


\subsection{Transfer Learning}
\label{sec:generalization}

To evaluate for cross-DLO transfer, we investigate the effort required to fine-tune the model, previously trained on the primary DLO (braided steel wire of \SI{1.5}{mm} diameter and \SI{550}{mm} length), to estimate shapes of new DLO types. We consider two new cables: a USB cable of diameter \SI{4.5}{mm} and length \SI{550}{mm}, and a braided steel wire of diameter \SI{2}{mm} and length \SI{750}{mm}. The latter DLO has the same material and structural properties as the primary DLO, but with slightly larger diameter and length. For each of the two cables, $50,000$ new data samples were collected along with a test set of 14,000 samples. Then, the model trained for the primary DLO was fine-tuned with a lower learning rate by an order of magnitude compared to the original training.

Figures \ref{fig:Transfer_USB}a and \ref{fig:Transfer_USB}b illustrate the scaling of RMSE and Chamfer errors relative to the size of the fine-tuning dataset, contrasted against models trained from scratch on the same novel data. In the zero-shot regime, the model exhibits significantly diminished performance, confirming that a network trained on a single DLO cannot inherently generalize to the distinct mechanical variations of novel DLOs without adaptation. However, the subsequent inclusion of limited data samples yields a decline in reconstruction error, demonstrating the sample efficiency of the fine-tuning process. Notably, models trained from scratch consistently yield inferior results compared to those initialized with the pre-trained weights. This performance gap suggests that the underlying physical representations learned from the primary DLO provide a robust structural prior that is highly transferable to DLOs with different mechanical properties. Shape estimation examples of the two DLOs can be seen in Figures \ref{fig:Cable_examples}b-c.

To evaluate scale sensitivity, we also tested the baseline model without retraining by shifting grasp positions to alter the effective length to $500\text{ mm}$ and $580\text{ mm}$. The mean reconstruction error increased to $36\text{ mm}$ and $40\text{ mm}$, respectively. This performance drop demonstrates that the model slightly struggles to generalize zero-shot to out-of-distribution scales, directly motivating future work on a length-aware architecture that ingests segment length as a dynamic input.

\subsection{Robustness to Partial Loss of Tactile Information}
\label{sec:tactile_ablation}

To evaluate the framework's resilience to partial data loss or hardware failures, we benchmarked performance under two paradigms where $\{1, 2, 3\}$ sensors are randomly removed: retraining the model on a reduced number of sensor, and 2) evaluating the accuracy of the model, without retraining, while masking faulty channels with blank inputs to simulate sudden runtime failure. As summarized in Table \ref{tab:tactile_ablation}, the retrained model exhibits excellent structural adaptation, maintaining a low RMSE even when relying on just half of the original tactile array. When subjected to sensor faults without retraining, performance degrades more rapidly due to the stark distribution shift at the input layer. This performance gap suggests a valuable future direction: incorporating stochastic sensor dropout during training to explicitly condition the network against runtime sensor faults.

\begin{table}[t]
  \centering
  \caption{Ablation study on the number of removed sensors}
  \label{tab:tactile_ablation}
  \begin{tabular}{l c c c}
    \toprule
    & Num. of Sensors & RMSE & Chamfer \\
      & removed & (mm) & dis. (mm) \\
    \midrule
    \multirow{3}{*}{\begin{tabular}[c]{@{}l@{}}New trained\\ model\end{tabular}} & 1 & $34.2\pm18.0$ & $30.1\pm14.2$ \\
    & 2 & $37.8\pm21.2$ & $34.0\pm16.9$ \\
    & 3 & $46.3\pm23.3$ & $39.6\pm17.6$ \\
    \midrule
    \multirow{3}{*}{\begin{tabular}[c]{@{}l@{}}Same model w/\\ fault sensors\end{tabular}} & 1 & $45.3\pm23.3$ & $36.4\pm16.7$ \\
    & 2 & $70.9\pm30.8$ & $53.4\pm20.2$ \\
    & 3 & $106.2\pm34.8$ & $76.9\pm21.5$ \\
    \bottomrule
  \end{tabular}
  \vspace{-0.5cm}
\end{table}

\section{Conclusions}

This paper demonstrates that high-fidelity global shape estimation of DLOs can be achieved using only tactile sensing and proprioception, bypassing visual limitations. By integrating high-resolution tactile contact gradients with global proprioceptive data through a physics-informed ViT framework, our approach accurately resolves high-dimensional DLO configurations. The framework effectively determines DLO-gripper orientation, maintains high performance in data-sparse regimes, and generalizes well to novel DLOs via sample-efficient transfer learning.


The performance degradation observed during sudden sensor failure and out-of-distribution states, such as near-buckling or rapid object slippage, motivates several directions for future work. To address these limitations, we intend to utilize high-fidelity simulation environments to generate diverse datasets across a wide range of mechanical properties and extreme deformation regimes, enabling a more generalized model that can handle sensor faults via dropout, as well as varying stiffness and friction coefficients. Additionally, we will explore incorporating the total wire length, measured directly by the robotic system, as a dynamic network input to enhance the model's adaptability to different DLO geometries without sacrificing tracking accuracy.


\bibliographystyle{IEEEtran}
\bibliography{Avishai_ref}

@article{Mishani2023,
title = {Learning configurations of wires for real-time shape estimation and manipulation planning},
journal = {Engineering Applications of Artificial Intelligence},
volume = {121},
pages = {105967},
year = {2023},
author = {Itamar Mishani and Avishai Sintov},
}

@Article{Bretl2014,
  Title                    = {Quasi-static Manipulation of a Kirchhoff Elastic Rod Based on a Geometric Analysis of Equilibrium Configurations},
  Author                   = {Bretl, Timothy and Mccarthy, Zoe},
  Journal                  = {Int. Jou. of Robotics Research},
  Year                     = {2014},
  Number                   = {1},
  Pages                    = {48--68},
  Volume                   = {33},
}

@Article{azulay2023allsight,
  author={Azulay, Osher and Curtis, Nimrod and Sokolovsky, Rotem and Levitski, Guy and Slomovik, Daniel and Lilling, Guy and Sintov, Avishai},
  journal={IEEE Robotics and Automation Letters}, 
  title={{AllSight}: A Low-Cost and High-Resolution Round Tactile Sensor With Zero-Shot Learning Capability}, 
  year={2024},
  volume={9},
  number={1},
  pages={483-490},
  doi={10.1109/LRA.2023.3333701}
}

@article{zhu_challenges_2021,
	author={Zhu, Jihong and Cherubini, Andrea and et al.},
  journal={IEEE Rob. \& Aut. Mag.}, 
  title={Challenges and Outlook in Robotic Manipulation of Deformable Objects}, 
  year={2022},
  volume={29},
  pages={67-77},
}

@article{tang_generative_2026,
	title = {Generative {3D} {State} {Estimation} for {DLOs} {From} {Partial} {Observations}},
	volume = {11},
	number = {1},
	journal = {IEEE Robotics and Automation Letters},
	author = {Tang, Yunxi and Yang, Tianqi and Huang, Jing and Chu, Xiangyu and Au, K. W. Samuel},
	month = jan,
	year = {2026},
	pages = {434--441},
}

@article{Caporali2026,
author = {Alessio Caporali and Ignacio Cuiral-Zueco and Gonzalo López-Nicolás and Gianluca Palli},
title ={Robotic perception and manipulation of deformable linear objects: A survey},
journal = {The International Journal of Robotics Research},
pages = {1-44},
year = {2026},
}

@article{zhaole_robust_2024,
	title = {A {Robust} {Deformable} {Linear} {Object} {Perception} {Pipeline} in {3D}: {From} {Segmentation} to {Reconstruction}},
	volume = {9},
	number = {1},
	urldate = {2026-02-01},
	journal = {IEEE Robotics and Automation Letters},
	author = {Zhaole, Sun and Zhou, Hang and Nanbo, Li and Chen, Longfei and Zhu, Jihong and Fisher, Robert B.},
	month = jan,
	year = {2024},
	pages = {843--850},
}

@inproceedings{monguzzi_tactile_2023,
	title = {Tactile based robotic skills for cable routing operations},
	urldate = {2026-02-01},
	booktitle = {{IEEE} {Int.} {Conf.} on {Robotics} and {Automation}},
	author = {Monguzzi, Andrea and Pelosi, Martina and Zanchettin, Andrea Maria and Rocco, Paolo},
	month = may,
	year = {2023},
	pages = {3793--3799},
}

@article{bergou_discrete_2008,
	title = {Discrete elastic rods},
	volume = {27},
	number = {3},
	urldate = {2026-01-22},
	journal = {ACM Tr. Graph.},
	author = {Bergou, Miklós and Wardetzky, Max and Robinson, Stephen and Audoly, Basile and Grinspun, Eitan},
	year = {2008},
	pages = {1--12},
}

@article{zhang_hardware_2022,
	title = {Hardware {Technology} of {Vision}-{Based} {Tactile} {Sensor}: {A} {Review}},
	volume = {22},
	number = {22},
	urldate = {2026-01-22},
	journal = {IEEE Sensors J.},
	author = {Zhang, Shixin and Chen, Zixi and Gao, Yuan and Wan, Weiwei and Shan, Jianhua and Xue, Hongxiang and Sun, Fuchun and Yang, Yiyong and Fang, Bin},
	year = {2022},
	pages = {21410--21427},
}

@article{she_cable_2021,
	author = {Yu She and Shaoxiong Wang and Siyuan Dong and Neha Sunil and Alberto Rodriguez and Edward Adelson},
    title ={Cable manipulation with a tactile-reactive gripper},
    journal = {The Int. Journal of Robotics Research},
    volume = {40},
    number = {12-14},
    pages = {1385-1401},
    year = {2021},

}

@article{sanchez_robotic_2018,
	author = {Jose Sanchez and Juan-Antonio Corrales and Belhassen-Chedli Bouzgarrou and Youcef Mezouar},
    title ={Robotic manipulation and sensing of deformable objects in domestic and industrial applications: a survey},
    
    journal = {The International Journal of Robotics Research},
    volume = {37},
    number = {7},
    pages = {688-716},
    year = {2018},
}

@inproceedings{yu_hand_2024,
	title = {In-{Hand} {Following} of {Deformable} {Linear} {Objects} {Using} {Dexterous} {Fingers} with {Tactile} {Sensing}},
	booktitle = {{IEEE}/{RSJ} {Int.} {Conf.} on {Intelligent} {Robots} and {Systems}},
	author = {Yu, Mingrui and Liang, Boyuan and Zhang, Xiang and Zhu, Xinghao and Sun, Lingfeng and Wang, Changhao and Song, Shiji and Li, Xiang and Tomizuka, Masayoshi},
	year = {2024},
	pages = {13518--13524},
}

@inproceedings{lv_learning_2023,
	author={Lv, Kangchen and Yu, Mingrui and et a.},
  booktitle={IEEE Int. Conf. on Robotics and Automation}, 
  title={Learning to Estimate 3-{D} States of Deformable Linear Objects from Single-Frame Occluded Point Clouds}, 
  year={2023},
  volume={},
  number={},
  pages={7119-7125},
}

@article{gu_learning_2025,
	title = {Learning {Graph} {Dynamics} {With} {Interaction} {Effects} {Propagation} for {Deformable} {Linear} {Objects} {Shape} {Control}},
	volume = {22},
	issn = {1558-3783},
	journal = {IEEE Trans. on Aut. Sci. and Eng.},
	author = {Gu, Feida and Sang, Hongrui and et al.},
    	year = {2025},
	pages = {10881--10892},
}

@article{xiang_trackdlo_2023,
	title = {{TrackDLO}: {Tracking} {Deformable} {Linear} {Objects} {Under} {Occlusion} {With} {Motion} {Coherence}},
	volume = {8},
	journal = {IEEE RA-L},
	author = {Xiang, Jingyi and Dinkel, Holly and Zhao, Harry and Gao, Naixiang and Coltin, Brian and Smith, Trey and Bretl, Timothy},
	year = {2023},
	pages = {6179--6186},
}

@article{yin_modeling_2021,
	title = {Modeling, learning, perception, and control methods for deformable object manipulation},
	volume = {6},
	number = {54},
	journal = {Science Robotics},
	author = {Yin, Hang and Varava, Anastasia and Kragic, Danica},
	year = {2021},
	pages = {eabd8803},
}

@inproceedings{tang_state_2017,
	title = {State estimation for deformable objects by point registration and dynamic simulation},
	booktitle = {{IEEE}/{RSJ} {Int.} {Conf.} on {Intel.} {Rob.} and {Sys.}},
	author = {Tang, Te and Fan, Yongxiang and Lin, Hsien-Chung and Tomizuka, Masayoshi},
	year = {2017},
	pages = {2427--2433},
}

@inproceedings{wilson_cable_2023,
	title = {Cable {Routing} and {Assembly} using {Tactile}-driven {Motion} {Primitives}},
	urldate = {2026-01-19},
	booktitle = {{IEEE} {Int.} {Conf.} on {Robotics} and {Automation}},
	author = {Wilson, Achu and Jiang, Helen and Lian, Wenzhao and Yuan, Wenzhen},
	year = {2023},
	pages = {10408--10414},
}

@article{mishani_real-time_2022,
	title = {Real-{Time} {Non}-{Visual} {Shape} {Estimation} and {Robotic} {Dual}-{Arm} {Manipulation} {Control} of an {Elastic} {Wire}},
	volume = {7},
	number = {1},
	journal = {IEEE Rob. and Auto. Letters},
	author = {Mishani, Itamar and Sintov, Avishai},
	year = {2022},
	pages = {422--429},
}

@article{noauthor_gelsight_nodate,
    AUTHOR = {Yuan, Wenzhen and Dong, Siyuan and Adelson, Edward H.},
    TITLE = {{GelSight}: High-Resolution Robot Tactile Sensors for Estimating Geometry and Force},
    JOURNAL = {Sensors},
    VOLUME = {17},
    YEAR = {2017},
    NUMBER = {12},
    ARTICLE-NUMBER = {2762},
}

@article{lambeta_digit_2020,
	title = {{DIGIT}: {A} {Novel} {Design} for a {Low}-{Cost} {Compact} {High}-{Resolution} {Tactile} {Sensor} {With} {Application} to {In}-{Hand} {Manipulation}},
	volume = {5},
	number = {3},
	journal = {IEEE Robotics and Automation Letters},
	author = {Lambeta, Mike and Chou, Po-Wei and Tian, Stephen and et al.},
	year = {2020},
	pages = {3838--3845},
}

@inproceedings{higuera_sparsh_2024,
	title = {Sparsh: {Self}-supervised touch representations for vision-based tactile sensing},
	author = {Higuera, Carolina and Sharma, Akash and Bodduluri, Chaithanya Krishna and Fan, Taosha and Lancaster, Patrick and Kalakrishnan, Mrinal and Kaess, Michael and Boots, Byron and Lambeta, Mike and Wu, Tingfan and Mukadam, Mustafa},
	year = {2024},
    booktitle = {Conf. on Robot Learning},
}

\end{document}